\documentclass[letterpaper]{article} 
\usepackage{aaai2026}  
\usepackage{times}  
\usepackage{helvet}  
\usepackage{courier}  
\usepackage[hyphens]{url}  
\usepackage{graphicx} 
\usepackage{natbib}  
\usepackage{caption} 
\usepackage{algorithm}
\usepackage{multirow}
\usepackage{booktabs}
\usepackage{array}

\usepackage{adjustbox} 
\usepackage{rotating}

\usepackage{algpseudocode}
\usepackage{amsmath}
\usepackage{amssymb}
\usepackage{subcaption}
\usepackage{bbding}
\usepackage{calc}
\usepackage{newfloat}
\usepackage{listings}
\DeclareCaptionStyle{ruled}{labelfont=normalfont,labelsep=colon,strut=off} 
\floatstyle{ruled}
\newfloat{listing}{tb}{lst}{}
\floatname{listing}{Listing}
\title{Enhanced Federated Deep Multi-View Clustering under Uncertainty Scenario}
\author{
    Bingjun Wei\textsuperscript{\rm 1,2}\equalcontrib,
    Xuemei Cao\textsuperscript{\rm 1,2}\equalcontrib,
    Jiafen Liu\textsuperscript{\rm 1,2}\thanks{Co-Corresponding author.},
    Haoyang Liang\textsuperscript{\rm 1,2}, 
    Xin Yang\textsuperscript{\rm 1,2}\footnotemark[2],
}

\affiliations{
    \textsuperscript{\rm 1}School of Computing and Artificial Intelligence, Southwestern University of Finance and Economics\\
    \textsuperscript{\rm 2}Cognitive Computing and Crowd Intelligence Lab, Southwestern University of Finance and Economics\\ 

    223081200006@smail.swufe.edu.cn, caoxuemei.qpz@gmail.com, jfliu@swufe.edu.cn, 223081200042@smail.swufe.edu.cn, yangxin@swufe.edu.cn
}

\usepackage{bibentry}

\begin{document}

\maketitle

\begin{abstract}
Traditional Federated Multi-View Clustering assumes uniform views across clients, yet practical deployments reveal heterogeneous view completeness with prevalent incomplete, redundant, or corrupted data. While recent approaches model view heterogeneity, they neglect semantic conflicts from dynamic view combinations, failing to address dual uncertainties: view uncertainty (semantic inconsistency from arbitrary view pairings) and aggregation uncertainty (divergent client updates with imbalanced contributions). To address these, we propose a novel Enhanced Federated Deep Multi-View Clustering framework: first align local semantics, hierarchical contrastive fusion within clients resolves view uncertainty by eliminating semantic conflicts; a view adaptive drift module mitigates aggregation uncertainty through global-local prototype contrast that dynamically corrects parameter deviations; and a balanced aggregation mechanism coordinates client updates. Experimental results demonstrate that EFDMVC achieves superior robustness against heterogeneous uncertain views across multiple benchmark datasets, consistently outperforming all state-of-the-art baselines in comprehensive evaluations.

\end{abstract}

\section{Introduction}

Multi-view Clustering (MVC) \cite{MVC,LIANG} improves clustering accuracy by fusing complementary and diverse information from heterogeneous views (e.g., images, texts, and videos). However, its centralized implementations require aggregating raw data into a central server, posing risks of data breaches \cite{data_silos} and violating privacy regulations \cite{fed_privacy}. Federated Learning (FL) \cite{survey1,survey2}, as a distributed collaborative framework, offers a new paradigm for decentralized processing of multi-view data, achieving privacy preservation through local model training and global parameter aggregation. 

\begin{figure}[htbp]
    \centering
    \includegraphics[width=\linewidth]{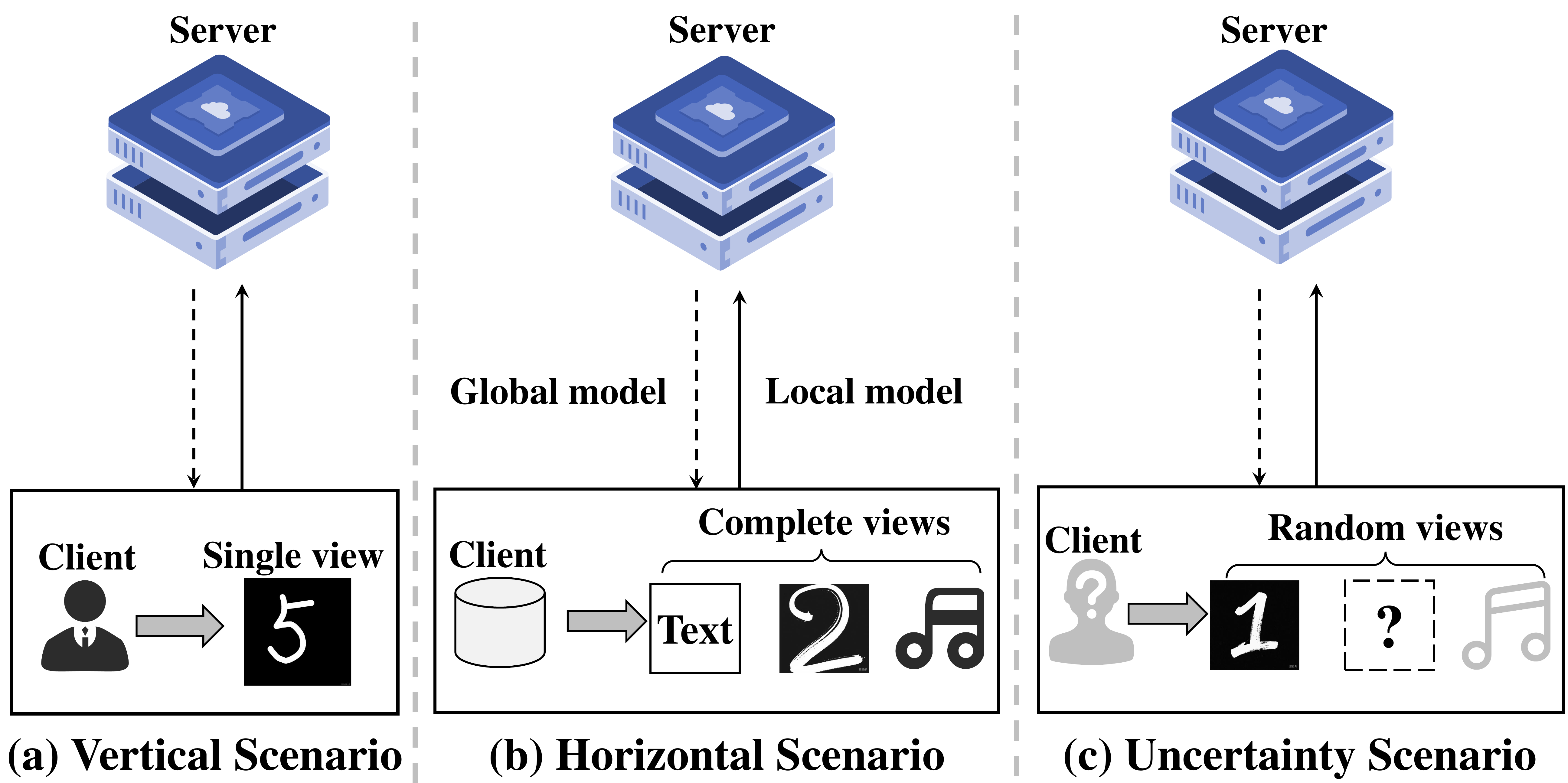} 
    \caption{Three scenarios of FedMVC}
    \label{setting}
\end{figure}

The combination of FL and MVC, termed Federated Multi-View Clustering (FedMVC) \cite{fedMVC_dl,TPAMI}, aims to explore more comprehensive clustering structures from unsupervised multi-view data distributed across multiple clients while preserving privacy. Current FedMVC methods seek to learn a global model from multi-view data distributed across multiple devices, integrating deep learning \cite{fedMVC_dl} and  matrix factorization \cite{matrix} to mine complementary information from views across clients. Through FL aggregation mechanisms \cite{fedavg}, these methods consolidate models on the server side to derive a robust global model.

\begin{figure*}[htbp]
    \centering
    \includegraphics[width=0.9\textwidth]{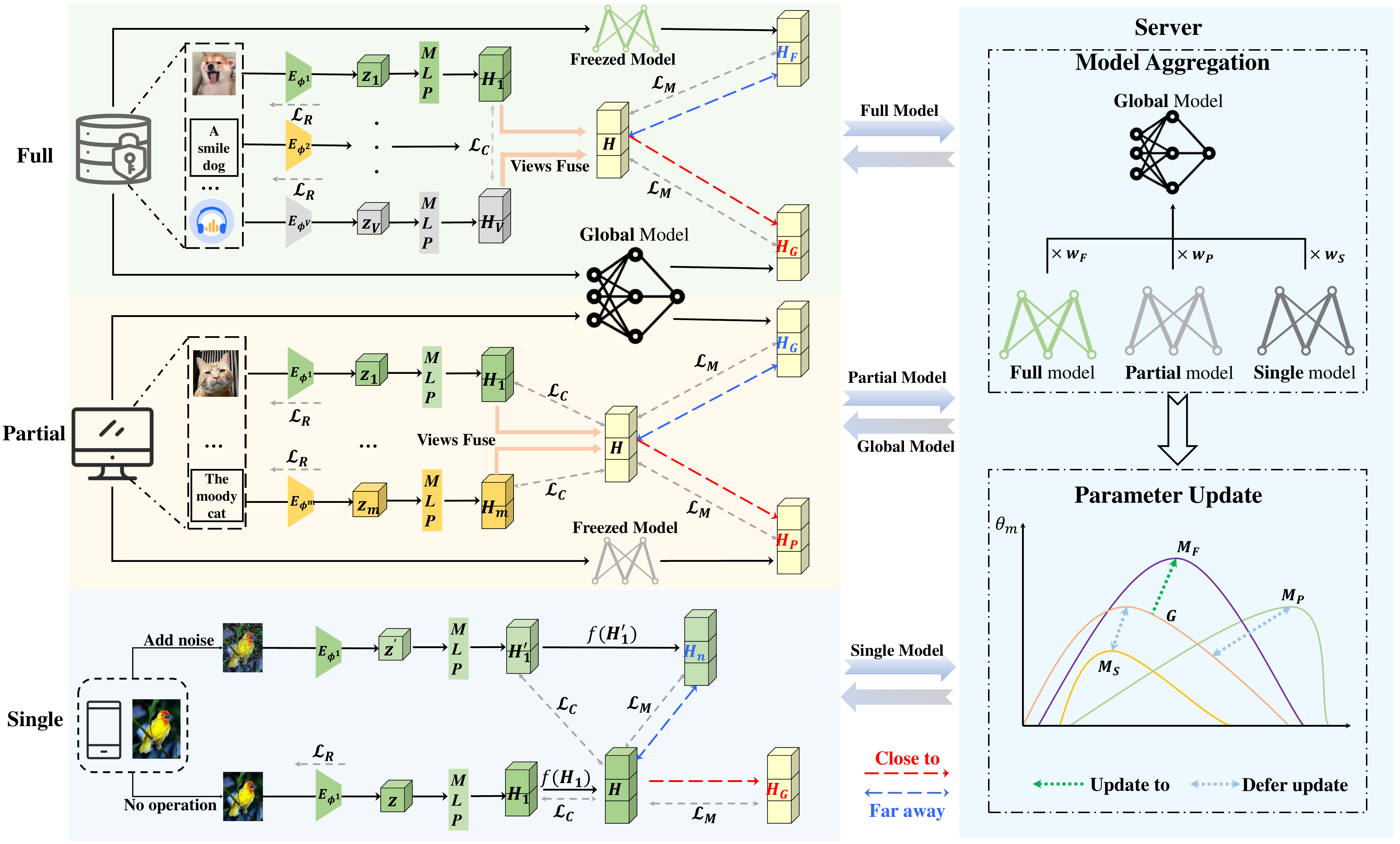} 
    \caption{In the EFDMVC approach, for each client, $\textbf{z}$ represents the learned feature representation, $\textbf{H}$ denotes the local semantic features of the client, $\textbf{H}_{G}$ stands for the global model features extracted based on client data, and the \textbf{Freezed Model} is preserved from the local model state saved at the end of the previous training round.}
    \label{overall}
\end{figure*}

As shown in Fig.~\ref{setting}, existing FedMVC methods generally presuppose a fixed view distribution across clients: either every client holds a single view~\cite{HFMVC} or all clients possess the complete set of views~\cite{H-FedMVC}. While such assumptions are workable under ideal conditions, they falter in real-world uncertain environments. FMCSC~\cite{FMCSC} does explore heterogeneous hybrid views, yet it is confined to the coexistence of the two static configurations above and cannot adapt to scenarios where the number of views per client changes dynamically, which severely limits its generalizability. In practice, clients may hold any number and combination of views: Hospital A has CT, MRI, and textual records simultaneously; Hospital B can only provide CT scans; Hospital C temporarily lacks MRI due to equipment maintenance and thus retains only CT and textual data. Such uncertain view distributions introduce multiple challenges:

$\textbf{\textit{View Uncertainty}}$: Clients hold varying view combinations. multi-view clients face feature conflicts from heterogeneous views; single-view clients suffer representation bias, degrading global clustering convergence.

$\textbf{\textit{Aggregation Uncertainty}}$: Inconsistent view structures across clients undermine the stability of federated optimization, making global model aggregation less robust.

To tackle the challenges outlined above, we propose Enhanced Federated Deep Multi-View Clustering (EFDMVC), as illustrated in Fig. \ref{overall}. EFDMVC is tailored for uncertain federated environments, where clients hold arbitrary subsets of views and face highly heterogeneous data distributions. We begin with feature initialization to filter out irrelevant information and align local semantic spaces. It then employs a hierarchical contrastive fusion strategy to alleviate the impact of view-specific noise and capture shared semantics across clients. Especially for single-view clients, we generate synthetic noisy samples to enable contrastive learning. To further enhance model consistency, we introduce a view adaptive drift mechanism that recalibrates update directions and mitigates model divergence. Additionally, we develop a view balanced aggregation strategy that dynamically adjusts client weights and harmonizes parameter update magnitudes across diverse view configurations, effectively reducing aggregation uncertainty. Our primary contributions are summarized as follows:

\begin{itemize}

\item We propose the EFDMVC method, which is the first time to model the challenge of the client dynamically holding arbitrary view subsets and highly heterogeneous data in the real scene.

\item We eliminated feature conflicts through hierarchical contrastive learning. And further propose an adaptive drift compensation module and adopt balanced aggregation to consider view quality and dynamically optimize the aggregation process.

\item Experimental analyses validate EFDMVC effectiveness, demonstrating superior clustering performance across various uncertainty scenarios.
\end{itemize}

\section{Related Work}

\subsection{Federated Multi-view Clustering}
As shown in Fig.~\ref{setting}, existing FedMVC methods are generally categorized based on data partitioning:

(1) Traditional FedMVC relies on conventional machine learning, employing static feature extraction and rigid aggregation rules. Fed-MVKM\cite{Fed-MVKM} alternately optimizes multi-view centroids between clients and server, while SFOMVC-TR\cite{SFOMVC-TR} innovatively applies Tucker decomposition to enable mixed client participation with factor matrix transmission.

(2) Vertical FedMVC assumes that $V$ views are distributed to $V$ single-view clients with shared samples. FedDMVC~\cite{FedDMVC} employs a global self-supervised contrastive framework for privacy-preserving view alignment, while HFMVC~\cite{HFMVC} enhances semantic consistency via a heterogeneity-aware dual contrastive mechanism.

(3) Horizontal FedMVC assumes multiple multi-view clients with non-overlapping samples. FMCSC~\cite{FMCSC} supports heterogeneous architectures with full-view and single-view clients by maximizing cross-client mutual information.

Despite their effectiveness, these methods struggle with more realistic scenarios involving uncertainty hybrid views, where the number and quality of views vary across clients. EFDMVC addresses this by bridging both view and aggregation uncertainty under a hybrid horizontal–vertical setting.

\subsection{Contrastive learning}

Contrastive Learning (CL) learns discriminative representations by constructing positive and negative pairs in unsupervised settings. Representative methods include SimCLR\cite{SimCLR}, which uses data augmentation and contrastive loss to structure the feature space; MoCo\cite{MoCo}, which introduces a momentum encoder and a negative queue to alleviate sample scarcity; and BYOL\cite{BYOL}, which eliminates explicit negatives by aligning online and target network outputs.

Integrating CL into Multi-View Clustering (MVC) and Federated Learning (FL) poses challenges in data heterogeneity and privacy. In MVC, methods like MFLVC\cite{MFLVC} leverage cross-view and label-level contrast to extract shared semantics while suppressing view-specific noise. In FL, MOON addresses model drift from $\textit{Non-IID}$ data via model-level contrast, while FedX\cite{MOON} employs cross-client knowledge distillation to jointly handle heterogeneity and privacy concerns.

\section{Method}
\subsection{Problem Setting}

In a uncertainty FedMVC environment, the \textbf{$C$} clients are categorized into three distinct types: single-view, partial-view, and full-view participants. Each client can have a maximum of $V$ views, and the number of non-overlapping samples per client can be approximated as $N$. The clients provide raw feature data, where multiple views share $K$ common clustering patterns to be discovered.

Each single-view client $\textbf{s}$ possesses a dataset $\mathcal{D}_{s} = \{\mathbf{x}_i^{v_s}\}_{i=1}^{N}$, where $v_s \in \{1,\ldots,V\}$ denotes the randomly assigned view type for that client. For partial-view client $\textbf{p}$, have dataset $\mathcal{D}_p = \{\mathbf{x}_i^{v \in \mathcal{V}_p}\}_{i=1}^{N}$ containing samples from a subset of views $\mathcal{V}_p \subset \{1,\ldots,V\}$ (where $|\mathcal{V}_p| \geq 2$). Full-view client $\textbf{f}$ maintain complete multi-view dataset $\mathcal{D}_f = \{(x_i^1, \ldots, x_i^V)\}_{i=1}^{N}$. The client model $M_C (\cdot)$ maps heterogeneous inputs to a unified space $\mathbb{R}^h$. After $\textbf{r}$ communication rounds, the server aggregates local models to produce global model $G(\cdot)$.

\subsection{Feature Alignment Initialization}
To address feature redundancy in heterogeneous multi-view data and cross-client view inconsistencies, we align semantic spaces across clients using autoencoders:

Each client processes its raw view data through view-specific encoders \cite{auto-encoder}, to obtain latent features $z_i^v = E_{\phi^v}(x_i^v) \in \mathbb{R}^{d_v}$ :
\begin{equation}
    \mathcal{L}_{R} = \frac{1}{N} \sum_{v \in \mathcal{V}'} \sum_{i=1}^{N} \|x_i^v - D_{\theta^v}(z_i^v)\|_2^2,
    \label{eq1}
\end{equation}
where $\mathcal{D}$ represents the local dataset ($\mathcal{D}_{f}$, $\mathcal{D}_{p}$, or $\mathcal{D}_{s}$), and $\mathcal{V}'$ denotes the set of available views for the client.

For each client, the features $\{\mathbf{z}^{v}\}_{v=1}^{|\mathcal{V}'|}$, obtained via Eq. (\ref{eq1}), contain a mixture of common semantics and view-private information. To extract higher-level representations, we treat $\{\mathbf{z}^{v}\}_{v=1}^{|\mathcal{V}'|}$ as low-level features and further process them using a feature $\textbf{MLP}$. This yields the high-level features $\{\mathbf{h}^{v}\}_{v=1}^{|\mathcal{V}'|}$, where each $\mathbf{h}^{v} \in \mathbb{R}^h$, we construct a non-linear mapping $\mathcal{H}(\textbf{Z}; \Psi): \mathbb{R}^{\sum_{v=1}^{|\mathcal{V}'|} d_v} \to \mathbb{R}^h$ defined as:  
\begin{equation}  
    \textbf{H} = \mathcal{H}(\textbf{Z}; \Psi) = \mathcal{H}\left(\textbf{Z}^1, \textbf{Z}^2, \ldots, \textbf{Z}^{|\mathcal{V}'|}; \Psi\right),  
    \label{eq2}  
\end{equation}  
where $\{\mathbf{h}_i\}_{i=1}^{N}=\mathbf{H}\in\mathbb{R}^{N\times d}$, and $\textbf{Z} \in \mathbb{R}^{N \times \sum_{v=1}^{|\mathcal{V}'|} d_v}$. Our objective is to preserve the discriminative power of the low-level features to avoid model collapse while learning the shared semantic representation $\mathbf{H}$ across views in the high-level feature space.

\subsection{Hierarchical Contrastive Fusion}

We propose a hierarchical contrastive learning fusion to address the issue of insufficient semantic consistency between different clients and adapt to heterogeneous view configurations. This contrastive framework enables each client to adverse effects of view-private information and learn  consistent semantics while preserving local privacy. Specifically, it operates at three granularity levels: 

\textbf{Full-view client}: It represents $\mathcal{V}' = \{1,\dots,V\}$ and $\mathcal{D} = \mathcal{D}_f$. Each high-level feature $\mathbf{h}_i$ has $(NV - 1)$ feature pairs, i.e. ,  $\{\mathbf{h}^v_i, \mathbf{h}^n_j\}_{\substack{n=1,\dots, V \\ j=1,\dots,N}}$, where 
$\{\mathbf{h}^v_i, \mathbf{h}^n_i\}_{v \neq n}$ are $(N - 1)$ positive feature pairs and 
the remaining $N(V - 1)$ feature pairs are negative feature pairs.

The similarities of positive pairs should be maximized, while those of negative pairs should be minimized. Inspired by NT-Xent \cite{SimCLR},  the cosine distance is applied to measure the similarity between two features:
\begin{equation}
d(\mathbf{a}, \mathbf{b}) = \frac{\langle \mathbf{a}, \mathbf{b} \rangle}{\|\mathbf{a}\| \|\mathbf{b}\|},
\label{eq3}
\end{equation}
 where $\langle\cdot,\cdot\rangle$ is dot product operator. Then, the feature contrastive loss between $\mathbf{h}^v_i$ and $\mathbf{h}^n_i$ is formulated as:
 \begin{equation}
 \mathcal{L}_C^{f}=-\frac{1}{N}\sum_{v=1}^{V}{{\sum_{i=1}^{N}\log\frac{e^{d(\mathbf{h}_{i}^{v},\mathbf{h}_{i}^{n})/\tau}}{\sum_{j=1}^{N}\sum_{v'=v,n}e^{d(\mathbf{h}_{i}^{v},\mathbf{h}_{j}^{v'})/\tau}-e^{1/\tau}}}},
 \label{eq4}
 \end{equation}
 where $\tau$ denotes the temperature parameter. 

Similar to learning the high-level features, we adopt CL to achieve this consistency objective. For the $v$-th view, the same cluster labels $\mathbf{Q}^v_{ j}={softmax}(\mathbf{h}^v_j)$ have $(VK - 1)$ label pairs, i.e., 
$\{\mathbf{Q}^v_{ j}, \mathbf{Q}^n_{ k}\}_{\substack{n=1,\dots,V \\ k=1,\dots,K}}$, where 
$\{\mathbf{Q}^v_{ j}, \mathbf{Q}^n_{ j}\}_{n \neq v}$ are constructed as $(V - 1)$ positive label pairs, and the remaining $V(K - 1)$ label pairs are negative label pairs. 

We further define the label contrastive loss between $\mathbf{Q}^v_i$ and $\mathbf{Q}^n_i$ as:
\begin{equation}
\begin{split}
\mathcal{L}_C^{l} = 
& -\frac{1}{K}\sum_{v=1}^{V}\sum_{j=1}^{K}\log\frac{e^{d(\mathbf{Q}_{ j}^{v},\mathbf{Q}_{ j}^{n})/\tau}}{\sum_{k=1}^{K}\sum_{v'=v,n}e^{d(\mathbf{Q}_{ j}^{v},\mathbf{Q}_{ k}^{v'})/\tau}-e^{1/\tau}} \\
& + \sum_{v=1}^V \sum_{j=1}^K s_j^v \log s_j^v,
\end{split}
\label{eq5}
\end{equation}
where $s_j^v=\frac{1}{N}\sum_{i=1}^{N}q_{ij}^v$, $q_{ij}^v$ represents the probability that the $i$-th sample belongs to the $j$-th cluster in the $v$-th view.

\textbf{Partial-view client}: $\mathcal{V}' = \mathcal{V}_p$ and $\mathcal{D} = \mathcal{D}_p$. It only contain a subset of the complete views, directly performing cross-view comparison faces the challenge of insufficient information. To address this, we propose a local contrastive learning module that aligns partial-view features $\mathbf{h}_{i}^{v}$ with common semantics $\mathbf{H}_{i}^p$. By establishing local-global correspondences, partial-view clients can leverage global common semantics to compensate for their perspective limitations, thereby obtaining more discriminative feature representations, the contrastive loss between $\mathbf{h}_{i}^{v}$ and $\mathbf{H}_{i}^p$ defined as:
\begin{equation}\mathcal{L}_{C}^{p}=-\frac{1}{N}\sum_{v=1}^{|\mathcal{V}_p|}\sum_{i=1}^{N}\log\frac{e^{d(\mathbf{H}_{i}^p,\mathbf{h}_{i}^{v})/\tau}}{\sum_{j\neq i}e^{d\left(\mathbf{H}_{i}^p,\mathbf{h}_{j}^{v}\right)/\tau}},
\label{eq6}
\end{equation}
where $\mathbf{h}_{i}^p$ is fused by the representations $\{\mathbf{h}_{i}^{v}\}_{v=1}^{|\mathcal{V}_p|}$.

\textbf{Single-view client}: $\mathcal{V}' = \{v_p\}$ and $\mathcal{D} = \mathcal{D}_s$. The absence of multi-view data necessitates an alternative contrastive module. We propose a noise-enhanced paradigm that establishes virtual negative pairs \cite{SimCLR} through perturbed samples while aligning local features with global prototypes. Specifically, we generate perturbed samples $\mathbf{h}^{'}_i$ by adding noise to local features $\mathbf{h}_i$, then contrast them with common semantics $\mathbf{H}_i^s$ using the following loss function:
\begin{equation}
\mathcal{L}_{{C}}^{s} = -\frac{1}{N}\sum_{i=1}^{N} \log \frac{e^{d(\mathbf{H}_i^s, \mathbf{h}_i)/\tau}}{e^{d(\mathbf{H}_i^s, \mathbf{h}_i)/\tau} +  e^{d(\mathbf{h}_i, \mathbf{h}^{'}_i)/\tau}},
\label{eq7}
\end{equation}
where $\mathbf{h}_i^s$ is obtained via a nonlinear transformation of the single-view representation $\mathbf{h}_i$, and $\mathbf{h}^{'}_i$ represents locally perturbed features generated through additive noise $\mathbf{h}^{'}_i = M_s({x}_i+ \epsilon_i) $ with $\epsilon_i \sim \mathcal{N}(0, \sigma^2\mathbf{I})$.

\subsection{View Adaptive Drift}

Inspired by MOON \cite{MOON,model-shift}, the model $M_f(\cdot)$ trained on the full-view client possesses the most data, while the globally aggregated model $G(\cdot)$ incorporates knowledge from all clients. we lveraging the global model broadcast at every communication round, and devise a view adaptive drift scheme: by carefully crafting diverse positive and negative sample pairs, we enable fine-grained exchange of client-specific knowledge, markedly suppressing semantic conflicts arising from missing views, while a regularization term steers the update direction to precisely mitigate aggregation uncertainty.

Therefore, to address the issue of view uncetainty among models, we employ CL to achieve specific model drifts for the three types of clients. The formulation is as follows: 
\begin{equation}
\mathcal{L}_{M} = -\log \frac{ e^{ d(\mathbf{H}, \mathbf{H}^+ / \tau ) }}{ e^{( d(\mathbf{H}, \mathbf{H}^+) / \tau ) }+ e^{(d(\mathbf{H}, \mathbf{H}^-)/ \tau )} }+\frac{\mu}{2}{||\omega_M-\omega_G||^2} ,
\label{eq8}
\end{equation}
where \(\mu\) controls the strength of the {proximal regularization term}, $\mathbf{H}^+$ represents the positive sample, and $\mathbf{H}^-$ represents the negative sample.  \(\omega_M\) and \(\omega_G\) are the {parameters of the local model} and {global model} , respectively.

In the $r$-th round of training, for the full-view client $\textbf{f}$, which is trained on complete multi-view data, its parameter update direction should be aligned with its own local model. Here, $\mathbf{H}^+$ is defined as the common semantics $\mathbf{\mathbf{H}}^f = M_f^{(r-1)}(\mathcal{D}_f)$, while $\mathbf{H}^-$ is given by the global model feature $\mathbf{H}^g = G^{(r-1)}(\mathcal{D}_f)$. For partial-view client $\textbf{p}$ and single-view client $\textbf{s}$, their local models are susceptible to data bias. The best available model they can access is $G^{(r-1)}(\cdot)$. Therefore, for these clients, $\mathbf{H}^+$ is set to $\mathbf{H}^g={G}^{(r-1)}({\mathcal{D}})$, while $H^-$ is set to their respective local models from the previous round, i.e., $\mathbf{H}^p = M_p^{(r-1)}(\mathcal{D}_p)$ for client $\textbf{p}$ and the noise fusion feature $\mathbf{H}' = M_s^{(r)}(\mathcal{D}_s')$ for client $\textbf{s}$, where $\mathcal{D}_s'$ represents the noise dataset. This module effectively addresses model drift for all three types of clients.

\subsection{Balanced View Aggregation}

Faced with the vast uncertainty of arbitrary view combinations in uncertain scenarios, conventional federated aggregation paradigms fall short; we must instead design an aggregation mechanism that simultaneously accounts for view complementarity and data trustworthiness, thereby achieving a stable and controllable global aggregation.

After completing local model training, each client uploads its trained model parameters to the central server. On the server side, a balanced view aggregation strategy is employed, which takes into account both the training sample size and multi-view data completeness of each participating node. This uncertainty weighted fusion mechanism ensures that the global model converges toward the optimal direction:
\begin{equation}
    w_k = \frac{\alpha_c \cdot n_c}{\sum_{i=1}^C (\alpha_i \cdot n_i)}, 
    \label{eq9}
\end{equation}
where $n_c$ denotes the sample size of client $c$ , and $\alpha_c$ prioritizes clients with richer view coverage. This design ensures that clients possessing both sufficient data and complete views exert greater influence during aggregation.

The global model $\omega_G$ is then updated by harmonizing all local models:
\begin{equation}
\omega_{G}^{(r)}= \sum_{c=1}^C w_c \cdot \omega_c^{(r)},
\label{eq10}
\end{equation}
where $\omega_c^{(r)}$ represents the parameters of client $c$ in the communication round $r$. By adaptively balancing data volume and view quality, this mechanism mitigates biases caused by skewed view distributions and drives the global model toward a Pareto-optimal state.

\begin{algorithm}
\caption{Pipeline of EFDMVC}
\label{alg:AFDMVC}
\begin{algorithmic}[1]
\Require Dataset with $N$ samples and $V$ views distributed among $C$ clients, with an expectation to be partitioned into $K$ clusters, with communication rounds $r$

\Ensure Global clustering predictions.

\For{each client $c \in C$}
    \State Consensus feature initialize by Eq.\eqref{eq1}.
\EndFor
\For{$r=1$ to  $R$}
    \For{$c = 1$ to $C$}
        \State Optimize the contrastive loss by Eq.\eqref{eq11} 
    \EndFor
    \State Aggregate client models by Eq.\eqref{eq8} and Eq.\eqref{eq9}.
    \State Distribute global model  to each client.    
\EndFor

\State Calculate the predictions by Eq.\eqref{eq12} and Eq.\eqref{eq13}.
\end{algorithmic}
\end{algorithm}

\subsection{Objective Function}
The respective total losses for full-views clients $f$ , partial-views clients $p$  and single-view clients $s$ :
\begin{equation}
\mathcal{L} = 
\begin{cases}
\mathcal{L}_f = \mathcal{L}_{R}^f + \alpha(\mathcal{L}_C^{l} + \mathcal{L}_C^{f}) + (1-\alpha)\mathcal{L}_M^{f} , \\
\mathcal{L}_p = \mathcal{L}_{R}^p + \alpha\mathcal{L}_C^{p} + (1-\alpha)\mathcal{L}_M^{p} , \\
\mathcal{L}_s = \mathcal{L}_{R}^s + \alpha\mathcal{L}_C^{s} + (1-\alpha)\mathcal{L}_M^{s} .
\end{cases}
\label{eq11}
\end{equation}

Where $\alpha$ denotes a hyperparameter that controls the balance between different objective components. In the optimization of EFDMVC, $\mathcal{L}_{R}$ serves as the foundation for learning distinct feature representations within individual clients. Concurrently, the cross-view consistency loss $\mathcal{L}_{C}$ and the model shift loss  and $\mathcal{L}_{M}$ are employed to discover common semantics across views operate synergistically to capture shared semantic patterns across multi-view data.

Finally, the server performs $K$-means on all common semantics $\mathbf{H}$ to obtain the global centroids $\mathbf{U}$: 
\begin{equation}
    \min_{\mathbf{U}_1, \mathbf{U}_2, \ldots, \mathbf{U}_K} \sum_{i=1}^N \sum_{j=1}^K \|\mathbf{H}_i - \mathbf{U}_j\|^2.
    \label{eq12}
\end{equation}

Therefore, the final prediction result for sample $i$ is:
\begin{equation}
    y_i = \arg\min_j \| \mathbf{H}_i - \mathbf{U}_j \|^2.
    \label{eq13}
\end{equation}

\section{Expriment}

\subsection{Experimental Settings}

\textbf{Scenario Configurations}: We implement a privacy-preserving FL framework with the following characteristics: (1) Multi-view data are preprocessed and randomly partitioned into training subsets following $\textit{IID}$ or $\textit{Non-IID}$ distributions; (2) Each client $\mathcal{C}_k$ receives uncertaintyally assigned views $|\mathcal{V}_k| \in [1,V]$ with corresponding data slices; (3) Local models are trained using client-specific multi-view data with encrypted parameter extraction; (4) The central server performs secure aggregation of uploaded parameters via cryptographic protocols; (5) Updated global models are redistributed for subsequent training rounds. The entire process enforces strict \emph{data localization} and \emph{client isolation} principles.

\textbf{Datasets}: We conducted experiments on six datasets: MNIST, HW, Multi-Fashion, BBC \cite{BBC}, UCI and  Caltech-5V. In order to better simulate real-world heterogeneous scenarios, the data partitioning is done using Dirichlet distribution \cite{Dir}. We configured four heterogeneous environments; Dirichlet (1), Dirichlet (10), Dirichlet (100), and $\textit{IID}$. Here, smaller values in Dirichlet $(\cdot)$  indicate greater heterogeneity for evaluating the fitness of the EFDMVC and the comparison methods.

\begin{table*}[ht]
\centering
\setlength{\tabcolsep}{5pt}{
\begin{tabular}{@{}m{0.2cm} | l | ccc | ccc | ccc | ccc@{}}
\toprule
\multirow{1}{*}{\rotatebox{90}{Data}} & 

\multirow{1}{*}{Heterogeneity} & 
\multicolumn{3}{c|}{Dirichlet (1.0)} & 
\multicolumn{3}{c|}{Dirichlet (10)} & 
\multicolumn{3}{c|}{Dirichlet (100)} & 
\multicolumn{3}{c}{$\textit{IID}$, Dirichlet ($\infty$)} \\
\cmidrule(lr){2-14}
 &Metrics & ACC & NMI & ARI & ACC & NMI & ARI & ACC & NMI & ARI & ACC & NMI & ARI \\
\midrule
\multirow{6}{*}{\rotatebox{90}{Fashion}} 
& MAGA              &15.15 & 0.00 & 0.00 
                  &13.44 & 0.00 & 0.00 
                  &11.22 & 0.00 & 0.00 
                  &10.76 & 0.00 & 0.00 \\				

& MFLVC  &45.22 &\underline{54.53} &37.08 
                    & \underline{69.28} & \underline{71.45} & \underline{58.22} 
                    & \underline{58.96}& \underline{67.74}& \underline{48.33}
                    &\underline{62.70} &68.60 & 50.23\\									 				

& SEM  &26.88 &25.40 & 10.04
                &26.83	&24.45& 8.82
                &24.47 &24.29& 8.51
                &24.9 & 25.90& 10.67\\

& HFMVC  & 28.01 & 21.96 & 10.49 
                    & 24.52 & 20.89 & 9.10 
                    & 21.81 & 14.07 & 6.99 
                    & 23.11 & 19.29 & 7.24  \\
& FMCSC   
                    & \underline{51.93} & 51.03 & \underline{36.28}
                    & 56.82 & 54.99 & 41.24 
                    & 58.21 & 60.71 & 46.46 
                    & 60.06 & \underline{63.96} & \underline{50.57}  \\
\cmidrule(lr){2-14}
& EFDMVC   
                    & \textbf{56.60} &\textbf{ 61.87 }& \textbf{46.00 }
                    & \textbf{78.03} & \textbf{78.01} & \textbf{69.50} 
                    & \textbf{85.16} & \textbf{77.82} & \textbf{72.88} 
                    & \textbf{86.89} & \textbf{78.75} & \textbf{74.95}  \\
\midrule
\multirow{6}{*}{\rotatebox{90}{MNIST }} 
& MAGA    
                    & 15.50 & 0.00 & 0.00   
                    & 12.45 & 0.00 & 0.00  
                    & 11.76 & 0.00 & 0.00   
                    & 10.89 & 0.00 & 0.00   \\
& MFLVC  
            & 15.86 & 8.18 & 1.34 
            & 53.70 & 44.95 & 28.38 
            & 39.62 & 31.67 & 8.10 
            & \underline{60.02} & \underline{47.77} & \underline{32.65}  \\

& SEM       

            & 23.19 & 15.66 & 7.11 
            & 20.01 & 17.14 & 6.34 
            & 20.86 & 15.57 & 6.16 
            & 18.94 & 14.81 & 4.82 \\
& HFMVC    
                        & 26.69 & 20.59 & 10.21 
                        & 20.98 & 10.31 & 4.03 
                        & 19.34 & 9.50 & 3.40 
                        & 21.57 & 17.33 & 7.71 \\

& FMCSC  
                    & \underline{41.44} & \underline{29.95} & \underline{20.06 }
                    & \underline{58.19} & \underline{39.81} & \underline{31.59} 
                    & \underline{55.11} & \underline{42.15} & \underline{32.58}
                    & 53.65 & 39.54 & 31.07 \\
\cmidrule(lr){2-14}
& EFDMVC             
                    & \textbf{50.67} & \textbf{36.58} & \textbf{27.13} 
                    & \textbf{61.40} & \textbf{45.93} & \textbf{38.81 }
                    & \textbf{61.94} & \textbf{48.90} & \textbf{42.44} 
                    & \textbf{75.50} & \textbf{57.80} & \textbf{54.5 }
                    \\

\midrule
\multirow{6}{*}{\rotatebox{90}{UCI }} 
& MAGA    
                    & 11.94 & 0.00 & 0.00 
                    & 11.39 & 0.00 & 0.00 
                    & 11.11 & 0.00 & 0.00 
                    & 10.83 & 0.00 & 0.00 \\
& MFLVC  & 28.10 & 33.91 & 13.73 
                    & 30.95 & \underline{37.60} & 20.06 
                    & \underline{34.25} & \underline{38.38} & \underline{23.56} 
                    & 36.90 & \underline{51.96} & \underline{33.80}  \\
& SEM     & 30.10 & 25.94 & 8.58 
                    & 30.45 & 27.49 & 11.55 
                    & 29.55 & 30.42 & 13.11 
                    & 35.05 & 31.67 & 14.35  \\
& HFMVC  & 29.70 & 28.56 & 12.56 
                    & 23.55 & 21.17 & 7.69 
                    & 23.05 & 25.02 & 10.96 
                    & 27.30 & 28.36 & 12.12 \\
& FMCSC  & \underline{48.90} & \underline{39.95} & \underline{27.40 }
                    & \underline{36.15} & {26.13} & \underline{14.54} 
                    & 31.50 & 21.39 & 12.40 
                    & \underline{38.40} & 31.62 & 18.73  \\
\cmidrule(lr){2-14}
& EFDMVC              & \textbf{52.05} & \textbf{47.28} & \textbf{33.13 }
                    & \textbf{63.60} & \textbf{55.11} & \textbf{44.08 }
                    & \textbf{55.95} & \textbf{52.19} & \textbf{39.08 }
                    & \textbf{55.45} & \textbf{51.02} & \textbf{37.09}  \\
\midrule
\multirow{6}{*}{\rotatebox{90}{HW }} 
& MAGA    & 11.94 & 0.00 & 0.00 
                    & 11.39 & 0.00 & 0.00 
                    & 11.11 & 0.00 & 0.00 
                    & 10.83 & 0.00 & 0.00  \\
& MFLVC  & \underline{34.60} & \underline{37.18} & 18.81 
                    & 37.20 & \underline{47.46} & \underline{26.87 }
                    & 34.45 & \underline{47.33} & \underline{28.07}
                    & \underline{36.40} & \underline{47.63} & \underline{30.08}\\
& SEM      & 29.05 & 32.92 & 13.02 
                    & 38.30 & 38.90 & 23.40 
                    & \underline{41.95} & 43.90 & 26.81 
                    & 32.25 & 36.12 & 17.18 \\
& HFMVC  & 31.30 & 32.83 & 17.45 
                    & 30.30 & 38.64 & 19.99 
                    & 30.25 & 37.12 & 18.03 
                    & 26.60 & 32.93 & 13.31 \\
& FMCSC  & \underline{34.60} & 34.40 & \underline{18.84 }
                    & \underline{38.20} & 34.39 & 19.73 
                    & 37.05 & 32.03 & 19.05 
                    & 35.70 & 31.13 & 16.47 \\
\cmidrule(lr){2-14}
& EFDMVC             & \textbf{46.80} & \textbf{47.38} & \textbf{30.18 }
                    & \textbf{61.95} & \textbf{54.00} & \textbf{40.26} 
                    & \textbf{58.55} & \textbf{54.23} & \textbf{41.02} 
                    & \textbf{60.70} & \textbf{56.92} & \textbf{44.47} \\
\midrule
\multirow{6}{*}{\rotatebox{90}{Cal-5V }} 
& MAGA    & 20.09 & 0.00 & 0.00 
                    & 17.56 & 0.00 & 0.00 
                    & 19.17 & 0.00 & 0.00 
                    & 20.00 & 0.00 & 0.00 \\
& MFLVC  & 31.64 & \underline{25.25} & \underline{14.65}
                    & 22.07 & 10.39 & 2.40 
                    & \underline{38.36} & \underline{31.04} & \textbf{19.19} 
                    & 34.07 & 22.57 & 11.62 \\
& SEM      & \textbf{39.07} & \textbf{27.28} & \textbf{16.46 }
                    & \underline{44.64} & \underline{34.10} & \underline{22.59 }
                    & 36.00 & \textbf{31.31} & 18.31 
                    & 33.29 & 27.07 & 14.57 \\
& HFMVC  & 33.21 & 22.09 & 9.98 
                    & 24.36 & 8.41 & 2.82 
                    & 23.00 & 9.68 & 3.19 
                    & 22.36 & 8.19 & 2.71 \\
& FMCSC & 34.36 & 19.92 & 13.28 
                    & 30.00 & 14.58 & 9.06 
                    & 31.93 & 17.79 & 12.05 
                    & \underline{44.79} & \underline{25.73} & \underline{21.31} \\
\cmidrule(lr){2-14}
& EFDMVC              & \underline{35.29} & 18.51 & 12.94 
                    & \textbf{52.07} & \textbf{39.98} & \textbf{30.93} 
                    & \textbf{42.64} & 29.13 & \underline{18.88 }
                    & \textbf{58.29} & \textbf{42.45} & \textbf{32.26} \\

\bottomrule
\end{tabular}}
\caption{Clustering performance. The mean values (\%) of 5 runs are reported. The best and the second best values are highlighted in \textbf{bold} and \underline{underline}.}
\label{tab:overall}
\end{table*}

\begin{figure*}[t]
    \centering
    \includegraphics[width=0.9\textwidth]{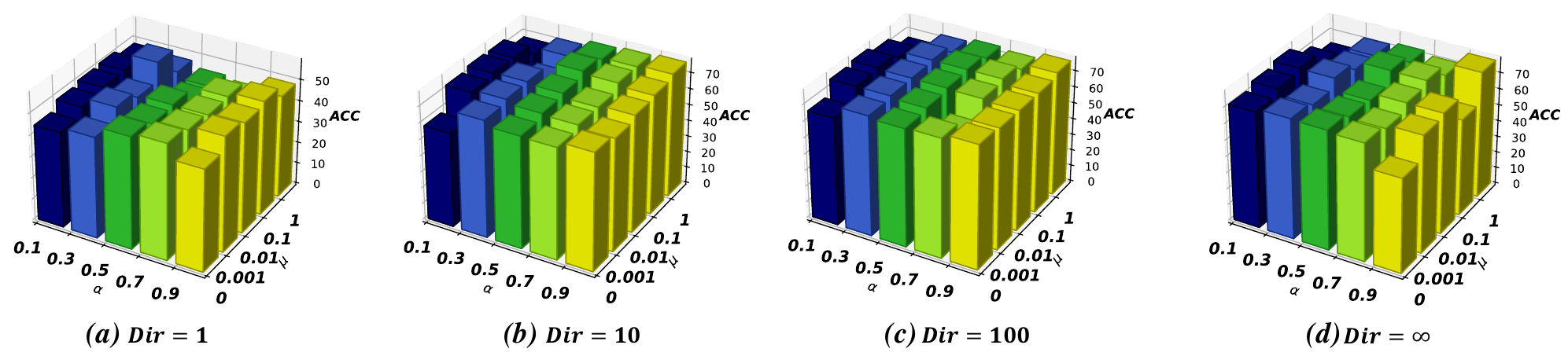} 
    \caption{Parameter comprison on different heterogeneous scenarios}
    \label{fig:four_images}
\end{figure*}

\textbf{Comparison Methods}: We select the following five state-of-the-art  methods for comparison: MFLVC \cite{MFLVC}, MAGA \cite{MAGA}, SEM \cite{SEM}, HFMVC \cite{HFMVC}, and FMCSC \cite{FMCSC}. It is worth noting that among the above methods, only HFMVC and FSCSC are applied in a federated environment, whereas the other methods are centralized methods. In order to ensure maximum fairness of comparison, we have made simple modifications to the alignment to support distributed environments. Given the scarcity of FedMVC research, these modifications were unavoidable.


\subsection{Result Comprison}

As systematically summarized in Table \ref{tab:overall} , we evaluate EFDMVC against state-of-the-art methods across diverse uncertainty FedMVC scenarios. The comparative analysis across four heterogeneous scenarios reveals four key findings:

(1) EFDMVC demonstrates remarkable clustering accuracy across all experimental configurations, particularly outperforming baseline methods that exhibit catastrophic performance degradation in such uncertainty scenarios. Notably on the Fashion dataset, EFDMVC achieves peak performance margins of 86.89\% (ACC), 78.75\% (NMI), and 74.95\% (ARI) , surpassing existing methods by significant margins.


(2) Compared to traditional centralized algorithms, EFDMVC achieves significant advantages. Furthermore, relative to decentralized algorithms like FMCSC and HFMVC, EFDMVC demonstrates superior performance through : adaptive parameter updating mechanisms, advanced multi-view pattern extraction, and effective resolution of cross-source heterogeneity challenges. Particularly in $\textit{IID}$ scenarios, the clustering performance of HFMVC and FMCSC nearly collapses (with ACC on HW being respectively only 25.60\% and 35.70\%, while EFDMVC achieves 60.70\%), demonstrating the practical feasibility of only EFDMVC.

(3) EFDMVC exhibits superior robustness. For instance, when the degree of heterogeneity ranges from Dirichlet (1.0) to $\textit{IID}$, EFDMVC's ACC on MNIST ranges from 50.67\% to 61.94\%, while on UCI, it ranges from 52.05\% to 55.45\%. In contrast, other methods exhibit much more significant fluctuations. Moreover, as the heterogeneity level decreases, the overall performance of EFDMVC tends to improve. In distributed scenarios, data is divided into more client nodes, resulting in fewer data points for each autoencoder to reconstruct. Conversely, as the heterogeneity level increases, the data reconstructed by individual autoencoders becomes more homogeneous, leading to better results.

\begin{table*}[ht]
\centering

\setlength{\tabcolsep}{2mm}
\renewcommand{\arraystretch}{1}
\begin{tabular}{|ccc|ccc|ccc|ccc|ccc|}
\hline
\multicolumn{3}{|c|}{Configuration} 
& \multicolumn{3}{c|}{\textbf{HW$_{Dir=1}$}} 
& \multicolumn{3}{c|}{\textbf{HW$_{Dir=\infty}$}} 
& \multicolumn{3}{c|}{\textbf{Caltech-5V$_{Dir=1}$}} 
& \multicolumn{3}{c|}{\textbf{Caltech-5V$_{Dir=\infty}$}} \\
\cline{1-15} 
$\mathcal{L}_C$ & $\mathcal{L}_M$ & Agg 
& ACC & NMI & ARI 
& ACC & NMI & ARI 
& ACC & NMI & ARI 
& ACC & NMI & ARI \\
\hline
 & \Checkmark & \Checkmark 
 & 27.8 & 18.8 & 9.1
 & 28.5 & 20.9 & 10.5
 & 29.6 & 10.2 & 6.7
 & 34.7 & 20.1 & 12.8 \\
\hline
\Checkmark & \Checkmark &  
 & 27.0 & 22.8 & 12.1 
 & 25.7 & 20.6 & 9.8
 & 23.0 & 7.3  & 3.4
 & 34.2 & 17.0 & 10.6 \\
\hline
\Checkmark &  & \Checkmark 
 & 46.0 & 42.5 & 30.4
 & 64.9 & 57.0 & 46.5 
 & 34.0 & 21.4 & 13.7 
 & 54.4 & 38.3 & 30.8 \\
\hline
\Checkmark & \Checkmark & \Checkmark 
 & \textbf{49.5} & \textbf{47.0} & \textbf{32.6}
 & \textbf{69.8} & \textbf{63.0} & \textbf{54.5}
 & \textbf{35.8} & \textbf{22.4} & \textbf{13.7}
 & \textbf{57.0} & \textbf{42.3} & \textbf{32.0} \\
\hline
\end{tabular}
\caption{Ablation studies on both $\textit{IID}$ and $\textit{Non-IID}$ settings}
\label{tab:performance}
\end{table*}

\subsection{Ablation Analysis}
We conduct ablation studies on HW and Caltech-5V under both $\textit{IID}$ and $\textit{Non-IID}$ scenarios to evaluate the impact of $\mathcal{L}_C$ (defined in Eq.\eqref{eq4} , Eq.\eqref{eq5}, Eq.\eqref{eq6} and  Eq.\eqref{eq7}), $\mathcal{L}_M$ (defined in Eq.\eqref{eq8}), and the aggregation strategy (defined in Eq.\eqref{eq9} and Eq.\eqref{eq10}), with results summarized in Table~\ref{tab:performance} :

(1) Removing $\mathcal{L}_C$ causes the ACC on {HW$_{Dir=1}$} to drop from {49.5\%} to {27.8\%} (21.7\% decline) and ARI from {32.6\%} to {9.1\%}, indicating that the absence of classification loss collapses semantic consistency and fails to capture cross-view clustering structures.
    
(2) Replacing our aggregation module with FedAvg drastically reduces ACC on {HW$_{Dir=\infty}$} from {69.8\%} to {25.7\%}, as FedAvg ignores view quality weights and fails to fuse heterogeneous client knowledge.

(3) Removing $\mathcal{L}_M$ has stronger impacts in $\textit{Non-IID}$ scenarios, e.g., ACC on {Caltech-5V$_{Dir=1}$} decreases by 12.8\% (vs. 5.6\% in $\textit{IID}$), due to aggravated local overfitting and failed global knowledge transfer.
    
(4) The full combination achieves peak performance (e.g., ACC at 69.8\% on {HW$_{Dir=\infty}$}) through complementary mechanisms: $\mathcal{L}_C$ enforces semantic supervision, $\mathcal{L}_M$ aligns models, and aggregation strategy uncertaintyally fuses heterogeneous knowledge.

\subsection{Parameter Analysis}

We conducted a detailed analysis of the two hyperparameters through a study on the MNIST dataset, as shown in Fig.\ref{fig:four_images}. The study focused on the impact of the loss balancing factor $\mathbf{\alpha}$ and the regularization coefficient $\mathbf{\mu}$. In the experiments, $\mathbf{\alpha}$ was varied across \([0.1, 0.3, 0.5, 0.7, 0.9]\), while $\mathbf{\mu}$ was varied across \([0, 0.001, 0.01, 0.1, 1]\).

The experimental results indicate that when $\mathbf{\alpha}$ is set to a small value (e.g., $ 0.1$), the clustering performance of EFDMVC significantly decreases. For instance, under the scenario of $\mathbf{\alpha}$, the ACC metric drops to its lowest value of 44.52\%. This may be attributed to the model overly relying on model shift while neglecting the contribution of local models in FL, thereby compromising the ability to extract consistent semantic representations from local multi-view data. In contrast, the regularization coefficient \(\mu\) exhibits relatively stable influence on the model's performance. Even with varying values of $\mathbf{\mu}$, EFDMVC maintains robust performance, indicating that the regularization term plays a crucial role in constraining parameter updates and further emphasizing the importance of the global model in FL frameworks. Considering the experimental results comprehensively, we selected $\mathbf{\alpha}=0.5$ and $\mathbf{\mu}=0.01$ as the default parameters to achieve optimal performance.

\subsection{Visualization Analysis}

\begin{figure}[t]
    \centering
    \includegraphics[width=\linewidth]{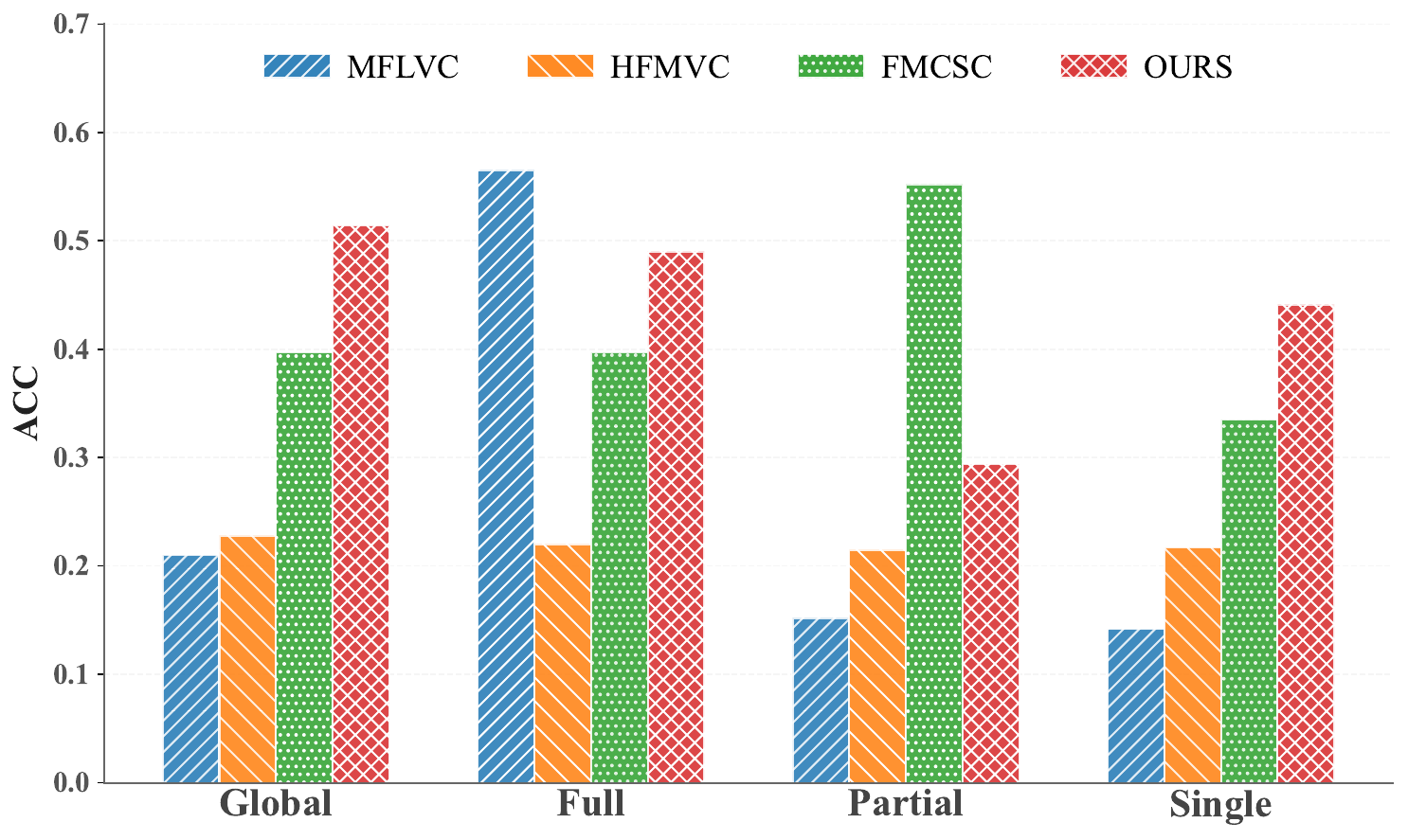} 
    \caption{Score comprison on different models}
    \label{score}
\end{figure}

We evaluate the clustering accuracy of different models on the Caltech-5V dataset under the $\textit{Non-IID}$ setting, as shown in Fig.~\ref{score}. Compared with the classical MFLVC, the horizontal FMCSC, and the vertical HFMVC, EFDMVC achieves consistently better performance across all three client models, and its aggregated global model significantly outperforms all baselines. These results demonstrate that the proposed balanced view aggregation module can effectively determine the aggregation direction, whereas existing methods fail to alleviate aggregation uncertainty and often suffer performance degradation after aggregation.
Furthermore, both the full-view and global models achieve better performance, validating our core hypothesis that they should guide the optimization of other models during model drift.

\section{Conclusion}
In this paper, we propose EFDMVC, which can handle practical uncertainty scenarios and explore data cluster structures distributed across multiple clients. First, we introduce consensus pre-training to align local models across all clients, eliminating view-specific representation biases while preserving task-relevant shared semantics. Then, we design local contrastive learning, model drift and global weighted aggregation to address view missing and $\textit{Non-IID}$ data distribution issues in client-side local training, effectively mitigating view imbalance interference on cluster boundaries. Extensive experiments demonstrate that EFDMVC outperforms SOTA methods in realistic and broad FedMVC scenarios. Although the current approach still has limitations in explicitly modeling inter-view semantic consistency. Future work will extend this framework to critical applications including uncertainty financial forecasting, while strengthening privacy preservation mechanisms and model lightweight deployment for real-world adaptability.

\bigskip

\bibliography{aaai2026}

\end{document}